\documentclass[sigconf]{acmart}
\usepackage{bm}
\usepackage{multirow}
\usepackage{soul}
\usepackage{makecell}

\setcopyright{acmlicensed}
\copyrightyear{2024}
\acmYear{2024}
\acmDOI{XXXXXXX.XXXXXXX}

\acmConference[KDD ’25]{the 31th ACM SIGKDD Conference on Knowledge Discovery and Data Mining }{August 3-7, 2025}{Toronto, ON, Canada}

\begin{document}

\title{Double-Path Adaptive-correlation Spatial-Temporal Inverted Transformer for Stock Time Series Forecasting}

\author{Wenbo Yan}
\affiliation{%
  \institution{School of Intelligence Science and Technology}
  \institution{Computational Intelligence Laboratory}
  \city{Peking University}
  \country{China}}
\email{wenboyan@stu.pku.edu.cn}

\author{Ying Tan}
\affiliation{%
  \institution{School of Intelligence Science and Technology}
  \institution{Institute for Artificial Intellignce}
  \institution{National Key Laboratory of General Artificial Intelligence}
  \institution{Key Laboratory of Machine Perceptron (MOE)}
  \city{Peking University}
  \country{China}}
\email{ytan@pku.edu.cn}

\begin{abstract}
    Spatial-temporal graph neural networks (STGNNs) have achieved significant success in various time series forecasting tasks. However, due to the lack of explicit and fixed spatial relationships in stock prediction tasks, many STGNNs fail to perform effectively in this domain. While some STGNNs learn spatial relationships from time series, they often lack comprehensiveness. Research indicates that modeling time series using feature changes as tokens reveals entirely different information compared to using time steps as tokens. To more comprehensively extract dynamic spatial information from stock data, we propose a Double-Path Adaptive-correlation Spatial-Temporal Inverted Transformer (DPA-STIFormer). DPA-STIFormer models each node via continuous changes in features as tokens and introduces a Double Direction Self-adaptation Fusion mechanism. This mechanism decomposes node encoding into temporal and feature representations, simultaneously extracting different spatial correlations from a double path approach, and proposes a Double-path gating mechanism to fuse these two types of correlation information. Experiments conducted on four stock market datasets demonstrate state-of-the-art results, validating the model's superior capability in uncovering latent temporal-correlation patterns.
\end{abstract}



\maketitle

\section{INTRODUCTION}
Time series data is ubiquitous in various aspects of life, including healthcare, transportation, weather, and finance, where a substantial amount of time series data is generated daily. If historical data can be utilized to uncover potential patterns and predict future changes, it could greatly enhance our quality of life. Initially, it was assumed that time series data is generated from certain processes, leading to the development of models such as \citet{box_jenkins_1970}, \citet{engle1982autoregressive} to approximate these processes. In recent years, numerous deep learning methods have been applied to the field of time series forecasting. To better model time series data, models based on recurrent neural networks (RNNs), such as \citet{qin2017dual}, as well as models based on convolutional neural networks (CNNs), such as \citet{dai2022price}, have been proposed. With the rise of the Transformer model in the field of natural language processing, Transformer-based time series forecasting models, such as \citet{crossformer}, have been introduced. As research has progressed, it has been found that for certain time series problems, such as traffic and climate forecasting, employing spatial-temporal prediction methods can better predict future changes. Consequently, many spatial-temporal graph neural networks (STGNNs) have been proposed and have achieved excellent results in fields like transportation. Moreover, to address the issue that Transformer-based methods cannot utilize spatial relationships, Spatial-Temporal Transformers (STTransformers) such as \citet{TFT}, \citet{STTN} have been proposed.

In fact, these time series modeling methods often fail to achieve satisfactory results in stock market prediction. Firstly, whether based on RNNs or temporal convolution methods, they typically focus on modeling the time series of individual nodes independently, without considering the interdependencies between different time series. However, in the stock market, the entire market functions as an integrated whole, and there are significant interactions between the time series of different stocks at any given moment. Therefore, ignoring the correlations between different nodes fails to effectively address the problem of stock prediction. Secondly, for Transformer-based methods, since the features of time series often lie in the variations between time steps and each time step itself contains low information content or even no actual meaning, using time steps as tokens to model time series with Transformers often results in poor performance.

While STGNNs can effectively utilize the correlations between nodes and integrate changes in other time series for comprehensive modeling, most spatial-temporal graph neural networks require pre-defined adjacency relationships, such as spatial distances in climate data or road connections in transportation networks. However, such relationships are not explicitly present in the stock market. Although ST-Transformers do not require predefined correlation relationships, existing methods still learn correlations between nodes from the perspective of time steps, and experiments show that the learned correlations are still limited. This issue is also present in STGNNs that do not require predefined adjacency matrices. In the latest research, some studies have recognized that time steps are not an ideal perspective for modeling time series. Crossformer uses both temporal and feature perspectives to model time series, while experiments with ITransformer indicate that using only features as tokens to model individual time series yields better results.

We believe that viewing time series from both feature and temporal perspectives reflects two completely different characteristics. While experiments with ITransformer have shown that modeling the time series of a single node using only features as tokens yields better results, when learning the correlations between different nodes using node time series, the feature and temporal perspectives can capture completely different node characteristics and yield two distinct types of correlations. Integrating these two different types of correlations can more comprehensively reflect the real influences between nodes. Based on this, we propose a novel method suitable for stock prediction, focusing on both the time series and the correlations between time series, named DPA-STIFormer.

Specifically, DPA-STIFormer consists of multiple Double-Path Adaptive-correlation Inverted Encoders and a single Decoder Block with Decomposed Fitting. The DPA-IEncoder includes an Inverted Temporal Block and a Double-Path Adaptive-correlation Block. The Inverted Temporal Block models the time series of each node using features as tokens. Considering that each feature has a different level of influence on the node, we introduce importance weights to ensure our method can account for the importance of tokens when calculating attention. The Double-Path Adaptive-correlation Block adaptively aggregates information from both the feature and temporal paths and learns two different types of correlations through the attention mechanism. We then propose a gating mechanism to effectively integrate information from the two paths. Following multiple encoder blocks, we introduce a specialized decoder suitable for stock prediction, which decomposes the prediction target into mean and deviation predictions, effectively handling both minor and drastic changes. In summary, the main contributions are as follows:

\begin{itemize}
\item We propose a novel Spatial-Temporal Transformer model, DPA-STIFormer, to address spatial-temporal prediction in stock markets. The model employs features as tokens and introduces an importance weight mechanism to better encode node characteristics.

\item We introduce DPABlock to model the correlation comprehensively from two perspectives. The Double Direction Self-adaptation Fusion mechanism is proposed to separate different features from time series. We learn the correlations through a double path approach, and a gating mechanism is proposed to integrate features from different perspectives.

\item We conducted extensive experiments on 4 stock datasets, and the experimental results demonstrate that our method can more effectively model the spatial-temporal correlations in time series.
\end{itemize}

\section{PRELIMINARY}
We first define the concepts of \textbf{Spatial-Temporal Forecasting Problem}. The spatial-temporal forecasting problem, refers to the task of using historical $T$ time steps data $\boldsymbol{X}\in{\mathbb{R}^{N \times T \times F}}$ and correlation adjacency matrix $\boldsymbol{A}\in{\mathbb{R}^{N \times N}}$ to predict future values for $t$ time steps $\boldsymbol{Y}\in{\mathbb{R}^{N \times t}}$.The adjacency matrix can be predefined or learned by the model, as in our approach. For each sample, there are $N$ nodes, and each node has a time series $X_i$, where $X_i$ contains $T$ time steps, and each time step has $F$ features. Additionally, the correlation adjacency matrix $A$ indicates the degree of correlation between the nodes, where $a_{ij}$ represents the correlation degree between node $i$ and node $j$. The neighbors set of node $i$ is represented as $K=\{j\ |\ j \neq i \ and\  a_{ij} \neq 0\}$.

\begin{figure*}
    \centering
    \includegraphics[width=\linewidth]{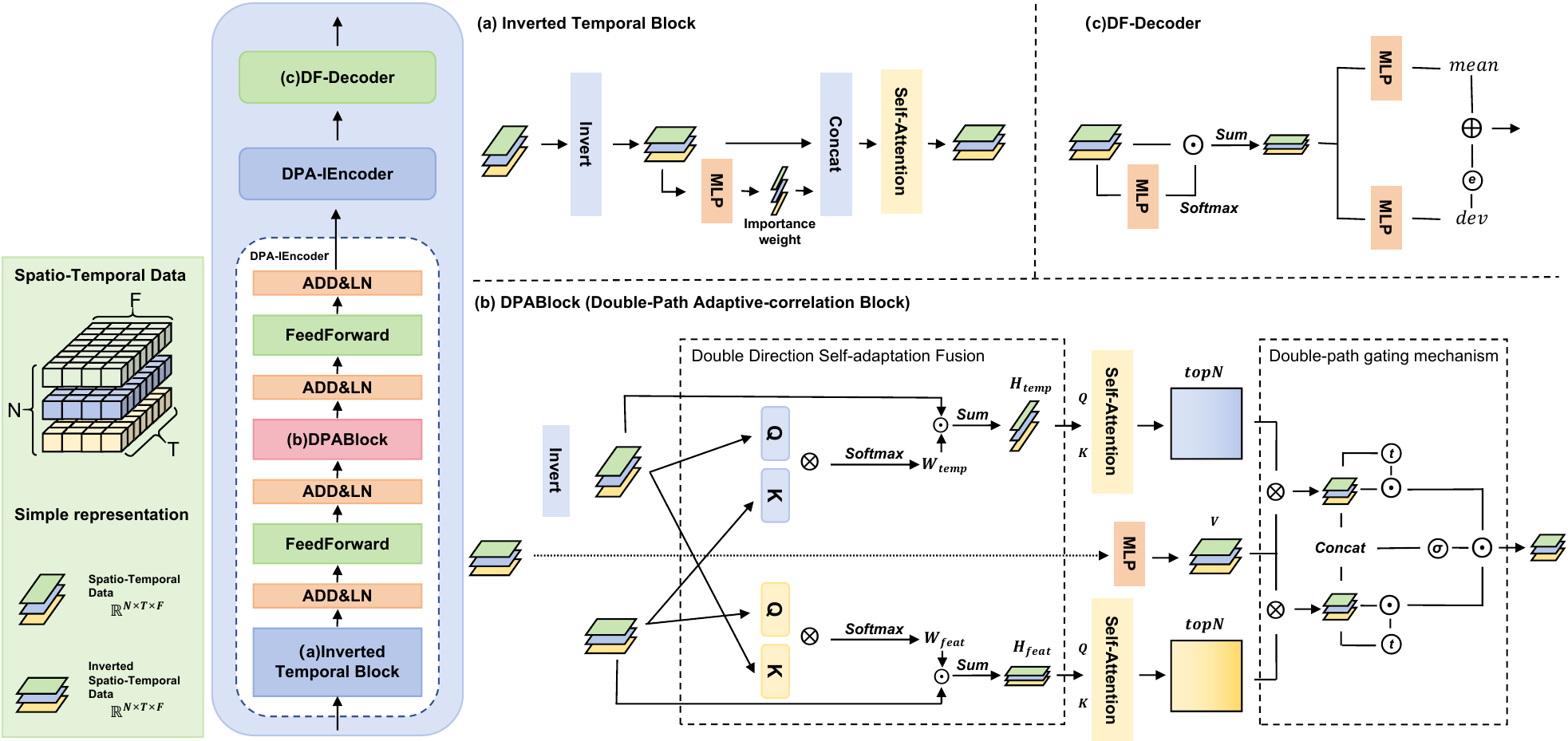}
    \caption{The overview of the proposed Double-Path Adaptive-correlation Spatial-Temporal Inverted Transformer.}
    \label{fig:arch}
\end{figure*}
\section{METHOD}

In this section, we present our approach. The precise architecture of the DPA-STIFormer is illustrated in Fig. \ref{fig:arch}.
\subsection{Overall}
The model is composed of multiple encoder blocks called Double-Path Adaptive-correlation Inverted Encoder (DPA-IEncoder) and a single Decoder Block with Decomposed Fitting (DF-Decoder) for stock prediction. Each DPA-IEncoder consists of an Inverted Temporal Block, a Double-Path Adaptive-correlation Block, and two feedforward neural networks, all interconnected by residual connections. The Inverted Temporal Encoder processes the time series $x_i$ for an individual node and the Double-Path Adaptive Correlation Encoder concurrently aggregates information from both temporal and feature dimensions, adaptively learns the relationships among neighboring nodes, and integrates the encoding results through a gating mechanism. Finally, the DF-Decoder demonstrates high capacity in stock prediction with proposed decomposed fitting.

\subsection{Inverted Temporal Block}
In many Transformer-based time series prediction approaches, features at the same time step are treated analogously to words in natural language processing. However, these methods often yield suboptimal performance. This is due to the inherently low information-to-noise ratio in time series data and the potential insignificance of individual time steps. Crucially, information is typically embedded in the temporal evolution of features rather than in isolated time points. Inspired by Itransformer\cite{itransformer} and crossformer\cite{crossformer} , we no longer consider features at the same time step as tokens, but rather the change of the same feature over time as a token. Thus, for the time series $x_i\in \mathbb{R}^{T\times F}$ in each node, We first Invert $x_i$ to get $x_i^{I}\in \mathbb{R}^{F\times T}$, rendering the temporal sequence of each feature as a token.
\begin{equation}
    x^{I}_{i}=invert(x_i), for\ i = 1,2,...,n
\end{equation}
where $invert(\cdot)$ stands for matrix transpose. Furthermore, considering the diverse features contained within the same node, it is evident that each feature exerts a different degree of influence on the node's variation, due to their distinct meanings. 
We introduce importance weights for each feature. Specifically, we pass $x_i^{I}$ through a two-layer fully-connected network to get their importance $w_i$ and  apply the softmax function to ensure that $\sum{w_i} =1$. Then, splice the $w_i$ to each token:
\begin{equation}
    w_i=softmax(W_2*\sigma(W_1*x^{I}_i+b_1)+b_2)
\end{equation}
\begin{equation}
    \hat{x}^{I}_i=<w_i,x^{I}_i>
\end{equation}
where $W_1$,$W_2$ represent learnable weight matrices, $b_1$, $b_2$ denote learnable bias vectors, $\sigma$ is the $ReLU()$ activation function, and $<\cdot>$ indicates matrix concatenation. The set of inverted time series with importance is denoted as $\hat{X}^{I}$.
\begin{equation}
    \hat{X}^{I}=[\hat{x}^{I}_1,\hat{x}^{I}_2,...,\hat{x}^{I}_n]
\end{equation}
Then, we model the time series of each node using the self-attention layer as in Transformer\cite{vaswani2017attention}.
\begin{equation}
     \mathcal{Q}= {W_Q^T} {\hat{X}^{I}}, \mathcal{K}= {W_K^T} {\hat{X}^{I}}, \mathcal{V}= {W_V^T} {\hat{X}^{I}}
\end{equation}
\begin{equation}
    {O^{I}}=softmax(\frac{ \mathcal{Q} \mathcal{K^T}}{\sqrt{d_k}}) \mathcal{V}
\end{equation}
where  $ {W_Q}$, $ {W_K}$, $ {W_V}$ are independent weights matrix that map $ {\hat{X}^{I}}$ to three different spaces, $O^{I}$ represents the output of the self-attention mechanism. Through this self-attention mechanism, we aim to learn the correlations between multiple features and develop better representations of each node. Subsequently, layer normalization and residual connections are applied to enhance the encoding process.
\begin{equation}
    \hat{X}^{I}_{out}=LN( O^{I}+residual(\hat{X}^{I}))
\end{equation}
where $LN(\cdot)$ stand for layer normalization, and $residual(\cdot)$ is a single-layer fully connection network to ensure equal dimension. Finally, the feedforward (FFD) neural network is applied:
\begin{equation}
    Z^{I}=LN( \hat{X}^{I}_{out}+FFD(\hat{X}^{I}_{out}))
\end{equation}

\subsection{Double-Path Adaptive-correlation Block}
Double-path Adaptive-correlation Block (DPABlock) is designed to adaptively model the correlations between nodes. It comprises three main components: Double Direction Self-adaptation Fusion, N-neighbor Adaptive Correlation Attention, and Double-Path Gating Mechanism.

\subsubsection{Double Direction Self-adaptation Fusion}
For spatial-temporal prediction problems, each node is often featured as a two-dimensional time series $x_i \in \mathbb{R}^{T\times F}$. The computational cost can become prohibitively large if all features are used to learn the correlations between nodes. A common approach is to use the features from the nearest time step or the mean of each time step as the node features. However, this approach has significant limitations. On the one hand, simple averaging over time steps does not provide a robust representation of the nodes. Different features hold varying levels of importance at different time steps, and averaging may dilute these distinctions, rendering the features less meaningful. On the other hand, many features of a time series are embedded in its successive changes, and aggregating multiple series into a single one may result in information loss compared to averaging over time steps.

We designed a double direction adaptive information fusion method as shown in Fig. \ref{fig:arch}. We first invert $Z^{I}$ again to obtain $Z$,
\begin{equation}
        Z=[z_{1},z_{2},...,z_{n}]
\end{equation}
Then, we respectively map $Z^{I}$ and $Z$ to different spaces through fully connected layers
\begin{equation}
\begin{split}
    \mathcal{Q}_{F} &= {W_{Q,F}^{T}} {Z}, \mathcal{K}_{F}= {W_{K,F}^{T}} {Z}\\
    \mathcal{Q}^{I}_{F} &= {W_{Q,F}^{I,T}} {Z^{I}}, \mathcal{K}^{I}_{F}= {W_{K,F}^{I,T}} {Z^{I}}
\end{split}
\end{equation}
where $\mathcal{Q}_{F}$,$\mathcal{K}_{F}$,$ \mathcal{Q}^{I}_{F}$, and $\mathcal{K}^{I}_{F}$ are the Query and Key of $Z$ and $Z^{I}$, respectively. They are all mapped to the same dimension. $W_{Q,F}^{T}$, $W_{K,F}^{T}$, $W_{Q,F}^{I,T}$ and ${W_{K,F}^{I,T}}$ are the corresponding weight matrices. We learn the weights of each feature at each time step by mutual querying, and we ensure that these weights sum to 1 in a certain direction using the softmax function.
\begin{equation}
\begin{split}
\mathcal{W}_{temp} &= softmax( \mathcal{Q}_{F}\mathcal{K}^{I}_{F})\\
        \mathcal{W}_{feat} &= softmax( \mathcal{Q}^{I}_{F}\mathcal{K}_{F})
\end{split}
\end{equation}
\begin{equation}
\begin{split}
    H_{temp} &= Sum(W_{temp}\cdot Z)\\
       H_{feat} &= Sum(W_{feat}\cdot Z^{I})\\
\end{split}
\end{equation}
where $\mathcal{W}_{temp} $ and $\mathcal{W}_{feat}$ are weighting matrices used to summarize information in the temporal and feature directions, $Sum(\cdot)$ is the summation of the last dimension, $H_{temp} \in \mathbb{R}^{N \times T} $ and $H_{feat} \in \mathbb{R}^{N \times F} $  are node representations that adaptively fusion from feature and temporal perspectives.  

\subsubsection{N-neighbor correlation attention}
Node representation $H_{feat}$ and $H_{temp}$ each capture different characteristics of the nodes. Therefore, we adaptively learn the correlations of the nodes from both perspectives. For each node and its neighbors, a close neighbor has a positive effect on the node, whereas information from more distant neighbors tends to be ineffective or even negative as the correlation diminishes. To address this, we propose an N-neighbor correlation attention mechanism. This mechanism learns correlations between nodes based on each node representation and preserves the top-n neighbors of each node through masking. We define the process $NcorrAttn$ as follows:
\begin{equation}
    \begin{split}
        &\mathcal{Q}_{G}= {W_{Q,G}^{T}} {H}, \mathcal{K}_{G}= {W_{K,G}^{T}} {H}, \mathcal{V}_{G}= {W_{V,G}^{T}} {Z^I}\\
        &NcorrAttn(H,X)=softmax(\frac{topN(\mathcal{Q}_{G}\mathcal{K}_{G}^{T})}{\sqrt{d_g}})\mathcal{V}_{G}
    \end{split}
\end{equation}
where ${W_{K,G}^{T}}$,${W_{Q,G}^{T}}$, ${W_{V,G}^{T}}$ are independent weights matrix, and $topN(\cdot)$ represents the retention of only the largest $N$ values in each row of the matrix, with $N$ typically set to 10\% of the number of nodes. Query $\mathcal{Q}_{G}$ and Key $\mathcal{K}_{G}$ are obtained by mapping node representation $H$ to different space, and Value $\mathcal{V}_{G}$ is mapped from $Z^I$. This approach is taken because the node representation captures only one aspect of the node's characteristics and is suitable for computing the N-neighbor attention matrix. However, relying solely on this enriched aspect can lead to the loss of the original features of the time series. Therefore, the Value should still be mapped from the original time series to retain the complete information. We can get two aspects of encoding:
\begin{equation}
    \begin{split}
        &O_{feat}= NcorrAttn(H_{feat},Z^{I})\\
        &O_{temp}= NcorrAttn(H_{temp},Z^{I})
    \end{split}
\end{equation}

Special Note that because $H_feat$, $H_temp$ are both from $Z^{I}$ and Value is also from $Z^{I}$, we use the same $\mathcal{V}_{G}= {W_{V,G}^{T}{Z^{I}}}$ as Value to save computational resources.
\subsubsection{Double-path gating mechanism}
In N-neighbor correlation attention, we model spatial relationships from both temporal and feature perspectives simultaneously, forming the integrated spatial-temporal encodings $O_{feat}$ and $O_{temp}$. For different spatial-temporal samples, the importance of the two paths may vary. Therefore, we propose a double-path gating (DPgate) mechanism to adaptively control the transmission and fusion of information from both sides. Our mechanism is illustrated in Fig. \ref{fig:arch}. DPgate mainly consists of two types of gates: self-passing gates and mutual-passing gates. Each path passes through a self-passing gate, which determines which information from that path passes through and which does not. Both paths simultaneously pass through a mutual-passing gate, where the proportion of information fusion is determined by the combined information from both paths. The specific calculation process is as follows:
\begin{equation}
    \begin{split}
        &g_{feat} = tanh(W_{s,feat}^TO_{feat}+b_{s,feat})\\
        &g_{temp} = tanh(W_{s,temp}^TO_{temp}+b_{s,temp})\\
        &m = \sigma(W_m^{T}\cdot[O_{feat},O_{temp}])\\
        &o_{feat}=g_{feat}*O_{feat}\\
        &o_{temp}=g_{temp}*O_{temp}\\
        &M=o_{feat}*m+o_{temp}*(1-m)
    \end{split}
\end{equation}
where $W_{s,feat}$, $W_{s,temp}$ are weight matrices of self-passing gate on each path, $W_m^{T}$ is the weight matrix of the mutual-passing gate, $tanh(\cdot)$ and $\sigma(\cdot)$ are the Tanh and Sigmoid activation functions, $g_{feat}$ and $g_{temp}$ are the self-passing gate on each path, $o_{feat}$ and $o_{temp}$ are the information passed through the self-passing gates, and $M$ is the information fused by the mutual-passing gate. The self-passing gate adjusts the effective information of each path itself, while the mutual-passing gate combines the information of both paths to control the passing ratio of the two paths.

\subsection{Decoder and decomposed fitting for stock prediction}
After processing through several Encoder Blocks, the model comprehensively extracts and integrates temporal and spatial information, represented as $M \in \mathbb{R}^{N \times F \times d_g}$. Given that we have inverted the input, the encoding process is no longer sequential. Consequently, we deviate from the traditional transformer decoding approach. Instead, we employ a Multilayer Perceptron (MLP) to map the encoded information directly to the prediction target. Initially, we integrate information from various features:
\begin{equation}
    \begin{split}
        &M_w = softmax(W_{1d}^TM)\\
        &M_m = Sum(M_w\cdot M)
    \end{split}
\end{equation}
where $W_{1d}$ is an $N \times 1$ mapping matrix. By applying the $softmax(\cdot)$ function and the $W_1d$, , each feature is assigned a weight that reflects its importance. Then aggregates all features to form a new mixed encoding $M_m \in \mathbb{R}^{N \times d_g}$.

For stock prediction problems, the mean of each stocks may be relatively stable over a period. Each time step can be seen as a deviation from this mean. We use  MLP to fit both the mean and deviation of each time series, and let the output be the sum of the mean and deviation.
\begin{equation}
    \begin{split}
        &mean = W_{mean}^TM_m+b_{mean}\\
        &dev = tanh(W_{dev}^TM_m+b_{dev})\\
        &\hat{y} = mean + e^{dev}
    \end{split}
\end{equation}
The prediction target is decomposed into an insensitive mean prediction and a sensitive deviation prediction. This approach enhances robustness and improves prediction performance. When each time step undergoes minor changes, the mean prediction remains relatively stable, while the sensitive deviation prediction provides an accurate fit. Conversely, when each time step experiences significant changes, the mean prediction adjusts accordingly, whereas the deviation prediction remains within a certain range, thereby preventing overfitting.

\subsubsection{Loss}

To attain high performance in time series prediction tasks, we consider both the accuracy of the predictions and the correlation between predictions across time steps in the loss function. Initially, we utilize Mean Squared Error loss $\mathcal{L}_{mse}$ as the loss function to evaluate the accuracy of the predictions and employ the Pearson correlation coefficient loss $\mathcal{L}_{pearson}$ to gauge the correlation of predictions within a time step.

The final form of the loss function is expressed as follows:
\begin{equation}\mathcal{L}=\lambda_{m}\mathcal{L}_{mse}+\mathcal{L}_{pearson}\end{equation}
where $\lambda_m$ denotes the weight of $\mathcal{L}_{mse}$ which is usually set to 0.1 to ensure that the two loss functions are of the same order of magnitude.

\begin{table}[htb]   
    \centering
    \small
    \setlength{\tabcolsep}{1mm}
{

    \begin{tabular}{c|c|c|c}
        \hline
        $\textbf{Dataset}$ & $\textbf{Samples}$ & $\textbf{Node}$ & \textbf{ Partition}\\
        \hline
        CSI500& 3159 & 500 & 2677/239/243\\ 
        CSI1000& 3159 & 1000 &  2677/239/243\\ 
        NASDAQ& 1225 & 1026 &  738/253/234\\ 
        NYSE& 1225 & 1737 &  738/253/234\\ 
        \hline
    \end{tabular}
    }
    \caption{The overall information for datasets}
    \label{tab:dataset}

\end{table}

\section{EXPERIMENTS}
In this section, we conduct experiments to demonstrate the effectiveness of our method. 
\subsection{Experimental Setup}

\subsubsection{Datasets}
We conducted experiments on four datasets: CSI500, CSI1000, NASDAQ, NYSE. Brief statistical information is listed in Table \ref{tab:dataset}. Detailed information about the datasets can be found in the appendix.
\subsubsection{Baseline}
We choose LSTM-based neural networks: FC-LSTM\cite{FCLSTM}, TPA-LSTM\cite{TPALSTM}, spatial-temporal graph neural networks: ASTGCN \cite{ASTGCN}, MSTGCN \cite{MSTGCN}, MTGNN \cite{MTGNN}, STEMGNN \cite{stemGNN}, STGCN \cite{STGCN}, DCRNN\cite{DCRNN}	 and transformer-based networks: Itransformer\cite{itransformer},Crossformer\cite{crossformer} as baseline to compare the performance of the models in all directions.

\subsubsection{Metrics}
We evaluate the performances of all baseline by nine metrics which are commonly used in stock prediction task including Information Coefficient (IC), Profits and Losses (PNL), Annual Return (A\_RET), Volatility (A\_VOL), Max Drawdown (MAXD), Sharpe Ratio (SHARPE), Win Rate (WINR), Profit/Loss Ratio (PL).

\begin{table*}[htbp]
  \centering


    \scriptsize
    
    \setlength{\tabcolsep}{1mm}
  {
  \scalebox{1}{
    \begin{tabular}{cc|cccccccc|cccccccc}
    \toprule
    \multicolumn{2}{c|}{\multirow{2}[4]{*}{}} & \multicolumn{8}{c|}{\textbf{CSI500}}                          & \multicolumn{8}{c}{\textbf{CSI1000}} \\
\cmidrule{3-18}    \multicolumn{2}{c|}{} & IC    & PNL   & A\_RET & A\_VOL & MAXD  & SHARPE & WINR  & PL    & IC    & PNL   & A\_RET & A\_VOL & MAXD  & SHARPE & WINR  & PL \\
    \midrule
    \multirow{6}[2]{*}{\textbf{STGNN}} & ASTGCN & 0.1022  & 0.3758  & 0.3727  & 0.0872  & 0.1813  & 4.2740  & 0.6281  & 2.0069  & 0.0840  & 0.7716  & 0.7652  & 0.1019  & 0.1285  & 6.8903  & 0.6694  & 3.0884  \\
          & STGCN & 0.0349  & 0.0466  & 0.0462  & 0.0676  & 0.1693  & 0.6841  & 0.5537  & 1.1236  & 0.0404  & 0.0388  & 0.0385  & 0.0865  & 0.2850  & 0.4449  & 0.5000  & 0.9310  \\
          & MSTGCN & 0.0732  & 0.0328  & 0.0325  & 0.0694  & 0.2246  & 0.4692  & 0.5793  & 1.9280  & 0.0919  & 0.8804  & 0.8732  & \textbf{0.1180 } & 0.1333  & 7.3996  & 0.6901  & 3.2232  \\
          & MTGNN & 0.1176  & 0.4652  & 0.4614  & 0.1006  & 0.1652  & 4.5845  & 0.6116  & 2.1265  & 0.0941  & 1.1608  & 1.1512  & 0.0756  & \textbf{0.0324 } & \textbf{15.2259 } & 0.7769  & \textbf{23.4614 } \\
          & STEMGNN & 0.0845  & 0.2124  & 0.2107  & 0.0842  & 0.1223  & 2.5016  & 0.5702  & 1.5455  & 0.0176  & 0.0154  & 0.0153  & 0.0205  & 0.0141  & 0.7448  & 0.0455  & 1.6529  \\
          & DCRNN & 0.0609  & 0.0899  & 0.0892  & 0.0839  & 0.2084  & 1.0630  & 0.5289  & 1.1794  & 0.0420  & 0.1080  & 0.1071  & 0.0850  & 0.2552  & 1.2591  & 0.4793  & 0.8216  \\
\cmidrule{1-2}    \multirow{2}[2]{*}{\textbf{RNN}} & TPA\_LSTM & 0.0052  & -0.0959  & -0.0951  & 0.0717  & 0.2364  & -1.3272  & 0.4959  & 0.7986  & 0.0344  & 0.0413  & 0.0409  & 0.0811  & 0.2304  & 0.5048  & 0.4959  & 0.9201  \\
          & FC\_LSTM & 0.1193  & 0.5402  & 0.5357  & \textbf{0.1056 } & 0.1428  & 5.0750  & 0.6033  & 2.3893  & 0.0482  & 1.0291  & 1.0206  & 0.1041  & 0.1053  & 9.8080  & 0.7190  & 5.4323  \\
\cmidrule{1-2}    \multirow{2}[2]{*}{\textbf{FORMER}} & CROSSFORMER & 0.0134  & 0.1447  & 0.1435  & 0.0779  & 0.2596  & 1.8423  & 0.4463  & 0.7295  & 0.0365  & 0.3537  & 0.3508  & 0.0964  & 0.2412  & 3.6383  & 0.5909  & 1.7650  \\
          & ITRANSFORMER & 0.0225  & -0.0517  & -0.0513  & 0.1031  & 0.2853  & -0.4976  & 0.5331  & 0.9226  & 0.0391  & 0.1947  & 0.1931  & 0.0945  & 0.2717  & 2.0267  & 0.5614  & 1.3757  \\
    \midrule
          & DPA-STIFormer    & \textbf{0.1312 } & \textbf{0.8794 } & \textbf{0.8721 } & 0.1033  & 0.1164  & \textbf{8.4395 } & \textbf{0.7273 } & \textbf{4.2453 } & \textbf{0.1404 } & \textbf{1.2809 } & \textbf{1.2703 } & 0.1111  & 0.0791  & 12.4611  & \textbf{0.8058 } & 9.2447  \\
    \midrule
    \multicolumn{2}{c|}{\multirow{2}[4]{*}{}} & \multicolumn{8}{c|}{\textbf{NYSE}}                            & \multicolumn{8}{c}{\textbf{NASDAQ}} \\
\cmidrule{3-18}    \multicolumn{2}{c|}{} & IC    & PNL   & A\_RET & A\_VOL & MAXD  & SHARPE & WINR  & PL    & IC    & PNL   & A\_RET & A\_VOL & MAXD  & SHARPE & WINR  & PL \\
    \midrule
    \multirow{6}[2]{*}{\textbf{STGNN}} & ASTGCN & 0.0400  & 0.0314  & 0.0322  & 0.3251  & 0.3936  & 0.9893  & 0.5171  & 0.8252  & 0.0323  & 0.2578  & 0.2644  & 0.1879  & 0.1369  & 1.4070  & 0.5128  & 1.2688  \\
          & STGCN & 0.0150  & 0.1009  & 0.1034  & 0.2707  & 0.1926  & 0.3821  & 0.5128  & 1.0667  & 0.0383  & 0.3145  & 0.3225  & 0.2408  & 0.1766  & 1.3396  & 0.5085  & 1.2675  \\
          & MSTGCN & 0.0351  & 0.0767  & 0.0786  & 0.2887  & 0.3035  & 0.2723  & 0.5085  & 1.0489  & 0.0227  & 0.2588  & 0.2655  & 0.1776  & \textbf{0.0973 } & 1.1524  & 0.4915  & \textbf{1.2845 } \\
          & MTGNN & -0.0035  & -0.0291  & -0.0299  & 0.1209  & \textbf{0.0604 } & -0.1000  & 0.4957  & 0.9831  & 0.0163  & 0.1375  & 0.1410  & 0.1537  & 0.1076  & 0.9179  & 0.5385  & 1.1803  \\
          & STEMGNN & 0.0254  & -0.1557  & -0.1597  & 0.2335  & 0.2691  & -0.6839  & 0.4915  & 0.8772  & 0.0171  & 0.0273  & 0.0280  & 0.2240  & 0.2597  & 0.1249  & 0.4786  & 1.0224  \\
          & DCRNN & 0.0016  & 0.0122  & 0.0125  & 0.1642  & 0.2281  & 0.0763  & 0.4701  & 1.0128  & -0.0056  & 0.0253  & 0.0259  & 0.1394  & 0.1273  & 0.1858  & 0.5284  & 1.0325  \\
\cmidrule{1-2}    \multirow{2}[2]{*}{\textbf{RNN}} & TPA\_LSTM & 0.0260  & 0.1729  & 0.1774  & 0.2557  & 0.3244  & 0.6937  & 0.5385  & \textbf{1.2037 } & 0.0270  & -0.0899  & -0.0922  & 0.2065  & 0.3087  & -0.4465  & 0.4786  & 0.9253  \\
          & FC\_LSTM & -0.0012  & 0.1255  & 0.1287  & \textbf{0.3391 } & 0.3694  & 0.3795  & 0.4872  & 1.0676  & 0.0083  & 0.2479  & 0.2542  & 0.2206  & 0.2393  & 1.2508  & 0.4829  & 1.2476  \\
\cmidrule{1-2}    \multirow{2}[2]{*}{\textbf{FORMER}} & CROSSFORMER & 0.0097  & 0.0204  & 0.0209  & 0.2767  & 0.2766  & 0.0756  & 0.5128  & 1.0128  & 0.0231  & 0.0611  & 0.0627  & 0.2081  & 0.1436  & 0.3010  & 0.4786  & 1.0544  \\
          & ITRANSFORMER & 0.0011  & -0.5296  & -0.5431  & 0.1141  & 0.5510  & -4.7592  & 0.4316  & 0.3288  & -0.0097  & -0.0931  & -0.0954  & 0.1158  & 0.1185  & -0.8241  & 0.5369  & 0.5229  \\
    \midrule
          & DPA-STIFormer    & \textbf{0.0453 } & \textbf{0.2135 } & \textbf{0.2190 } & 0.2987  & 0.1999  & \textbf{1.8115 } & \textbf{0.5427 } & 1.1256  & \textbf{0.0416 } & \textbf{0.3902 } & \textbf{0.4002 } & \textbf{0.3199 } & 0.1462  & \textbf{1.4945 } & \textbf{0.5470 } & 1.2320  \\
    \bottomrule
    \end{tabular}%
    }
  }
  \label{tab:main}%
    \caption{Comparison results on four datasets. $\downarrow$ indicates that the smaller the metric is better. The best result is in bold.}
\end{table*}%

\subsection{Main Results}
We compare our proposed model DPA-STIFormer, with 10 baseline models, encompassing three common types of models in time series forecasting: RNNs, Transformers, and STGNNs. The comparison results are presented in Table 2. Overall, our model achieves the best results across all four datasets. Our model consistently attains optimal results on multiple metrics simultaneously, and its performance improves with larger datasets. Compared to STGNNs such as ASTGCN, our model improves IC by an average of 30\%, A\_RET by an average of 20\%, and WINR by more than 3\% on average. The significant improvement is attributed to our model's ability to comprehensively capture the correlations between node time series. Many spatial-temporal graph neural networks rely on predefined graphs, or the relationships between nodes learned by these models are not sufficiently comprehensive. This also underscores the effectiveness of our DPABlock.

Compared to RNNs, our model demonstrates superior performance, particularly in terms of IC. On average, our model achieves more than a 30\% improvement in both PNL and A\_RET. Additionally, SHARPE shows significant improvement in the CSI500, NASDAQ, and NYSE datasets, ranking second only to MTGNN in the CSI1000 dataset. In comparison to Transformer-based models, our model also outperforms significantly. Transformers models, exhibit much lower IC values than our model and STGNNs. Metrics such as WINR and A\_RET further highlight that former-based models perform poorly in stock prediction without this crucial node correlation information. This also indicates the ineffectiveness of using time as tokens for modeling time series, aligning with the findings of \citet{itransformer}. Our model, on the other hand, uses features as tokens to model time series and employs a double-path approach to comprehensively capture the correlations between nodes. Using a gating mechanism, it effectively integrates information from both node correlations and time series. The experimental results validate the effectiveness of our method.

\begin{table}[htbp]
  \centering
    \small
    \setlength{\tabcolsep}{1mm}
{
    \begin{tabular}{c|ccc}
    \toprule
          & IC    & A\_RET & SHARPE \\
    \midrule
     DPA-STIFormer    & 0.1312  & 0.8721  & 8.4395  \\
    w/o DPgate & 0.1170  & 0.4804  & 4.4671  \\
    w/o temporal path & 0.1135  & 0.6465  & 6.7517  \\
    w/o feature path & 0.1190  & 0.5154  & 6.1406  \\
    w/o Inverted Temporal Block & 0.1010  & 0.5819  & 5.6339  \\
    w/o importance weigh & 0.1100  & 0.6142  & 5.0505  \\
    \bottomrule
    \end{tabular}%
}
  \label{tab:ablation}%
    \caption{Ablation Study}
\end{table}%

\subsection{Ablation Study}
To validate the effectiveness of the key components, we conduct an ablation study on the CSI500 dataset, comparing the changes in IC, A\_RET, and SHARPE. Specifically, we tested the performance impact of removing the following components: DPgate, temporal path, feature path, Inverted Temporal Block, and importance weight. Firstly, removing any path from the double-path structure or the final gating mechanism during data fusion results in approximately a 15\% decrease in IC, with varying degrees of decrease in A\_RET and SHARPE. The most severe drop occurs when the gating mechanism is removed, resulting in about a 25\% decrease. This indicates that removing any path leads to biased learning of correlations. Removing the gating mechanism and merging the information with equal weights leads to rough information fusion, causing adverse effects. From the changes in A\_RET and SHARPE, it is clear that the information learned from the feature path has a greater impact on the results. 

Compared to the model without the Inverted Temporal Block, removing this block results in a significant decrease in IC and SHARPE. This indicates a decline in overall predictive performance and increased instability, as the absence of individual denoising and modeling of each time series negatively affects the modeling of correlations. For the model without importance weight, the results show that including feature importance enhances overall predictive capability and returns, enabling the model to better utilize different features to model node characteristics. However, the decrease in SHARPE suggests that the addition of importance weight also leads to an increase in risk.

\begin{figure}[thbp]
    \centering
    \includegraphics[width=\linewidth]{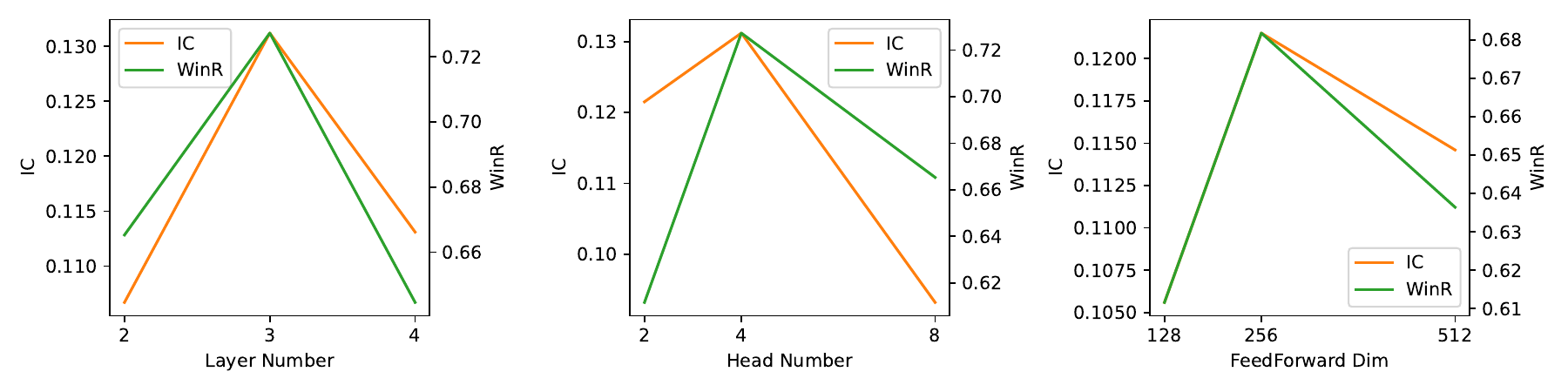}
    \caption[width=\textwidth]{Hyper-parameter Study}
    \label{fig:params}
\end{figure}

\subsection{Hyper-parameter Study}

We conducted extensive parameter experiments on the CSI500 dataset to examine the effects of Layer Number $n_l$, the number of attention heads $n_h$, and the dimensions of the feedforward neural network $d_{model}$. For $n_l$, we found that the model performs best with 3 layers. When the number of layers is less than 3, the IC is relatively low, indicating that the model suffers from overfitting due to insufficient parameters, leading to inadequate overall predictive capability. When the model has more than 3 layers, there is a significant decrease in WINR, likely because the data is insufficient to support training, resulting in underfitting. Since each additional layer significantly increases the model parameters, only 1 or 2 layers are needed for smaller datasets. Regarding $n_h$,, the optimal performance was observed with 4 heads. With fewer heads, the model cannot effectively capture all relationships. Conversely, with too many heads, the model tends to overfit, diminishing its overall predictive capability. Comparing different $d_{model}$ , we see that smaller dimensions greatly impact model performance, because each token cannot be fully encoded and distinguished. However, when the $d_{model}$ is too large, the model's parameter count increases rapidly, leading to gradual underfitting.
\subsection{Visualization}
\subsubsection{Double-path Visualization}
\begin{figure}[thbp]
    \centering
    \includegraphics[width=\linewidth]{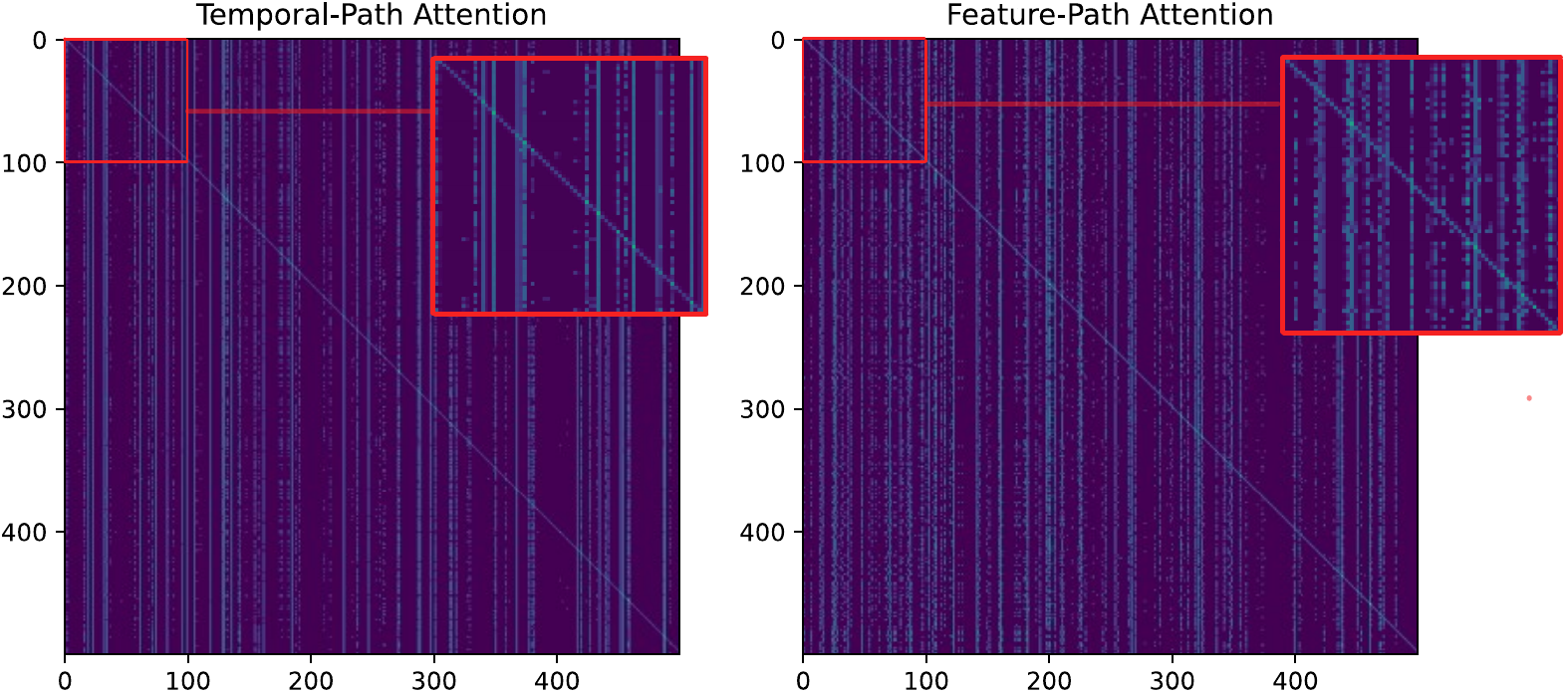}
    \caption[width=\textwidth]{Visualization of Double-path Attention}
    \label{fig:attn_vir}
\end{figure}
To deeply analyze the Double-path adaptive correlation block, we visualized the attention computation results of the final layer of the Double-path attention mechanism. It can be observed that the attention matrix obtained from the N-neighbor correlation attention is sparse, with each node focusing on its nearest N nodes. By comparing the two attention matrices, we can find that the Feature-path and Temporal-path have learned two different types of adjacency relationships. In the Temporal-path Attention, there are certain leading nodes that influence all other nodes. Conversely, in the Feature-path Attention, the adjacency relationships between nodes are more complex, with the N-neighbors of different nodes being more diversified.

\subsubsection{Mean-deviation Prediction Visualization}

\begin{figure}[thbp]
    \centering
    \includegraphics[width=\linewidth]{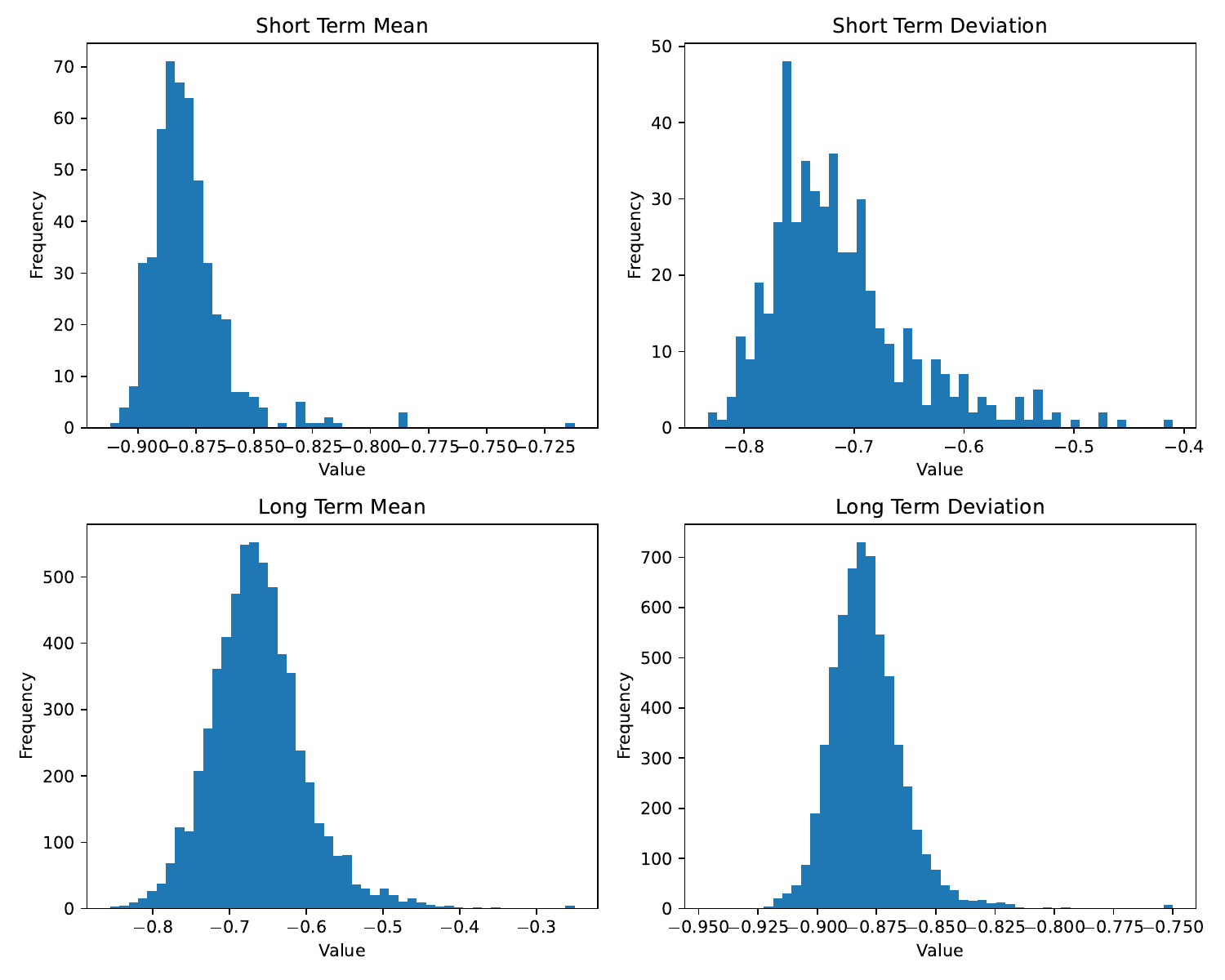}
    \caption[width=\textwidth]{Visualization of Mean-deviation Prediction}
    \label{fig:pred_vir}
\end{figure}
We visualize the distributions of mean and deviation for the short and long term to better illustrate the advantages of our Mean-deviation Prediction. The short-term refers to a period of continuous time-step prediction, and it can be seen that the distribution of the mean is denser while the distribution of the deviation is sparser, which indicates that in continuous prediction, the mean of our prediction method varies slowly, fitting the variation of the time series through the deviation. Long-term refers to the prediction after an interval of several time steps, where the range of the mean value has changed significantly while the deviation is still concentrated in a smaller range. This indicates that after an interval of time steps, Mean-deviation Prediction relies on the mean value to fit the drastic change in the time series and avoids too large a change in the deviation.

\section{RELATED WORK}
\subsection{Spatial-Temporal Graph Neural Networks}
In recent years, there has been a burgeoning interest in integrating temporal and spatial information within time series data, leading to the development of a series of models known as Spatial-Temporal Graph Neural Networks. STGCN \cite{STGCN} was the first to introduce Graph Convolutional Neural Networks (GCNs) \cite{GCN} into time series forecasting. Subsequently, additional convolution-based models, such as Graph Wave Net \cite{Gwave}, MT-GNN \cite{MTGNN}, StemGNN \cite{stemGNN}, H-STGCN \cite{H-STGCN}, and GSTNet \cite{GSTNet}, have been proposed. These models incorporate various gating mechanisms atop convolutions to enhance feature capture. Concurrently, other studies have focused on more complex convolutional structures, such as ST-GDN [31] and ST-ResNet \cite{ST-GDN}, which achieve improved performance through sophisticated architectural designs and mechanisms.

Moreover, some works, including ARGCN \cite{ARGCN}, DCRNN \cite{DCRNN}, and TGCN \cite{TGCN}, combine Recurrent Neural Networks (RNNs) with Graph Neural Networks (GNNs).
Additionally, with the advent of Transformers, many models have integrated transformer architectures or attention mechanisms into spatial-temporal modeling, as evidenced by ASTGCN \cite{ASTGCN}, STGNN \cite{sTGNN}, and GMAN \cite{GMAN}.
\subsection{Spatial-Temporal Transformer}
Recent advancements in spatial-temporal modeling have seen the integration of Transformers to capture complex dependencies across time and space. Notable models such as the TFT \cite{TFT} and the STTN \cite{STTN} have demonstrated the efficacy of combining temporal and spatial attention mechanisms. These models leverage the self-attention mechanism to effectively handle the intricacies of spatial-temporal data, thereby improving forecasting accuracy and interpretability.

\section{CONCLUSION}

In this paper, we  proposes a novel DPA-STIFormer. The Inverted Temporal Block models the temporal sequences of each node by treating features as tokens and introduces importance weights to enable the attention mechanism to consider the significance of features. The Double-Path Adaptive-correlation Block is introduced to model the correlations between nodes. The Double Direction Self-adaptation Fusion adaptively blends temporal and feature perspectives from the node embeddings, modeling inter-node correlations from both paths simultaneously. Finally, the proposed Double-path Gating Mechanism integrates the encodings from both paths. Extensive experiments on four real-world stock datasets, along with visualization of the learned correlations, demonstrate the superiority of our approach.

\bibliographystyle{ACM-Reference-Format}
\bibliography{acmart}

\clearpage
\appendix
\section{Dataset}
\begin{itemize}
    \item CSI500:  CSI500 is a stock dataset that contains the performance of 500 small and medium-sized companies listed on the Shanghai and Shenzhen stock exchanges. It contains daily frequency data for 500 stocks, with a total time step of 3159 and a feature number of 45.
    \item CSI1000: CSI1000 contains daily frequency data for 1000 stocks, with a total time step of 3159 and a feature number of 45.
    \item NASDAQ: NASDAQ contains daily frequency data for 1026 stocks in National Association of Securities Dealers Automated Quotations , with a total time step of 1225 and a feature number of 5.
    \item NYSE: NYSE contains daily frequency data for 1737 stocks in New York Stock Exchange, with a total time step of 1225 and a feature number of 5.
\end{itemize}
\subsection{Model Implementation}
We set the time window to 30, meaning each moment looks back at the past 30 time steps. The dimensions of the feedforward neural network and all attention mechanisms are set to 256. The number of attention heads is set to 4. For the CSI500 and CSI1000 datasets, 3 layers of DPA-IEncoder are used. For the NASDAQ and NYSE datasets, which have less data, only 1 and 2 layers of DPA-IEncoder are used.

\section{Experimental environment}
We ran the experiment on a server containing 4 NVIDIA GeForce RTX 3090 and AMD EPYC 7282 16-Core Processor, repeating each experiment five times and averaging the results.

\section{Metrics}
\begin{itemize}
    \item IC: the Pearson correlation between predicted ranking scores and real ranking scores, widely used to evaluate the performance of stock ranking.
    $$IC = \frac{{\text{Cov}(\hat{Y},Y)}}{{\sigma_{\hat{Y}} \sigma_Y}}$$
    \item PNL: the aggregate profits and losses of trading strategies, i.e., the cumulative returns.
    $$PNL= \sum_{n=1}^{N_d}{r_n}$$
    \item AR:  the annualized returns of trading strategies, denoting the profitability of portfolios. In the experiments, the number of trading days in one year is assigned as 240.
    $$AR=\frac{240}{N_d} \sum_{n=1}^{N_d}{r_n}\times 100\%$$
    \item VOL: the annualized volatility of portfolio returns, indicating the stability of portfolios.
    $$VOL=\sigma(R)\times\sqrt{240}\times 100\%$$
    \item MDD: a commonly used metric to measure risk control ability, which is the largest decline in portfolios from peak to trough.  
    $$MDD= {r_{max}}-{r_{min}} \times 100\%$$
    \item Sharpe: a widely-used metric to measure the investment quality of portfolios, which is the ratio of the average and standard deviation of portfolio returns, considering both of profitability and investment risk.
    $$Sharpe = \frac{\mu(R)}{\sigma(R)} \times \sqrt{240}$$ 
    where $R=(r_1,r_2,...,r_(N_d)$ is the portfolio returns, and the annualized values of SR are used in the experiments.
    \item Calmar: another metric to measure the quality of portfolios, which is the ratio of average returns and Max Drawdown (MDD).
    $$Calmar = \frac{\mu(R)}{MDD}$$
    \item WinR: a widely-used metric to evaluate the accuracy of the predictions, which is the ratio of the number of positive purchases to the total number of purchases. 
    $$WinR= \frac{N_{+}}{N_{total}}\times 100\%$$
    \item PL-ratio: the profit/loss ratio is the average profit on winning trades divided by the average loss on losing trades over a specified time period.
    $$PL-ratio = \frac{\sum{r_{+}}}{N_{win}} \bigg/ \frac{\sum{r_{-}}}{N_{lose}} $$

\end{itemize}

\subsection{Losses}
MSE Loss:
$$ \mathcal{L}_{mse} =\frac{1}{N}\sum_{i=1}^{N}{(\widehat{y_i}-y_i)^2}$$
Pearson Correlation Coefficient Loss:
$$ \mathcal{L}_{pearson} =-\frac{\sum_{i=1}^{N}(y_i-\overline{y})(\widehat{y_i}-\overline{\widehat{y}})}{\sqrt{\sum_{i=1}^{N}(y_i-\overline{y})^2}\sqrt{\sum_{i=1}^{N}(\widehat{y_i}-\overline{\widehat{y}})^2}}
$$
\end{document}